\pdfoutput=1

\documentclass[11pt]{article}

\usepackage[preprint]{acl}

\usepackage{times}
\usepackage{latexsym}

\usepackage[T1]{fontenc}

\usepackage[utf8]{inputenc}

\usepackage{microtype}

\usepackage{inconsolata}

\usepackage{graphicx}

\title{\ours: Improving Video-Text Retrieval Through Relevance-Based Augmentation Using Large Foundation Models}
\usepackage{listings}
\usepackage{multirow}
\usepackage{multicol}
\usepackage{microtype}
\usepackage{bbding}
\usepackage{bm}
\usepackage{colortbl}

\usepackage{amsmath}
\usepackage{amssymb}
\usepackage{amsfonts}
\usepackage{booktabs}
\usepackage{algorithm}
\usepackage{algorithmic}
\usepackage{natbib}
\usepackage{cleveref} 
\usepackage{subcaption}
\usepackage{supertabular}
\usepackage{longtable}  
\usepackage{tabularray}  
\usepackage[utf8]{inputenc}
\UseTblrLibrary{booktabs}

\usepackage[american]{babel}

\usepackage{color}
\usepackage{graphicx}
\allowdisplaybreaks
\usepackage{cleveref}

\newcommand{\eg}{\textit{e.g.}}
\newcommand{\ie}{\textit{i.e.}}

\usepackage{booktabs}
\usepackage{multirow}
\usepackage{graphicx}

\newcommand{\ours}{\textsc{dReAm}}

\author{Yimu Wang$^{1}$, Shuai Yuan$^{2}$, Bo Xue$^{3}$, Xiangru Jian$^{1}$, Wei Pang$^{1}$, Mushi Wang$^{1}$, Ning Yu$^{4}$\\
$^{1}$ University of Waterloo, $^{2}$ Duke University\\ $^{3}$ City University of Hong Kong, $^{4}$ Netflix Eyeline Studios \\
$^{1}$ \{yimu.wang,xiangru.jian,w3pang,m358wang\}@uwaterloo.ca\\
$^{2}$ shuai@cs.duke.edu, $^{3}$ boxue4-c@my.cityu.edu.hk, $^{4}$ ningyu.hust@gmail.com
}

\begin{document}
\maketitle

\begin{figure*}[t!]
\centering
\includegraphics[width=1\textwidth]{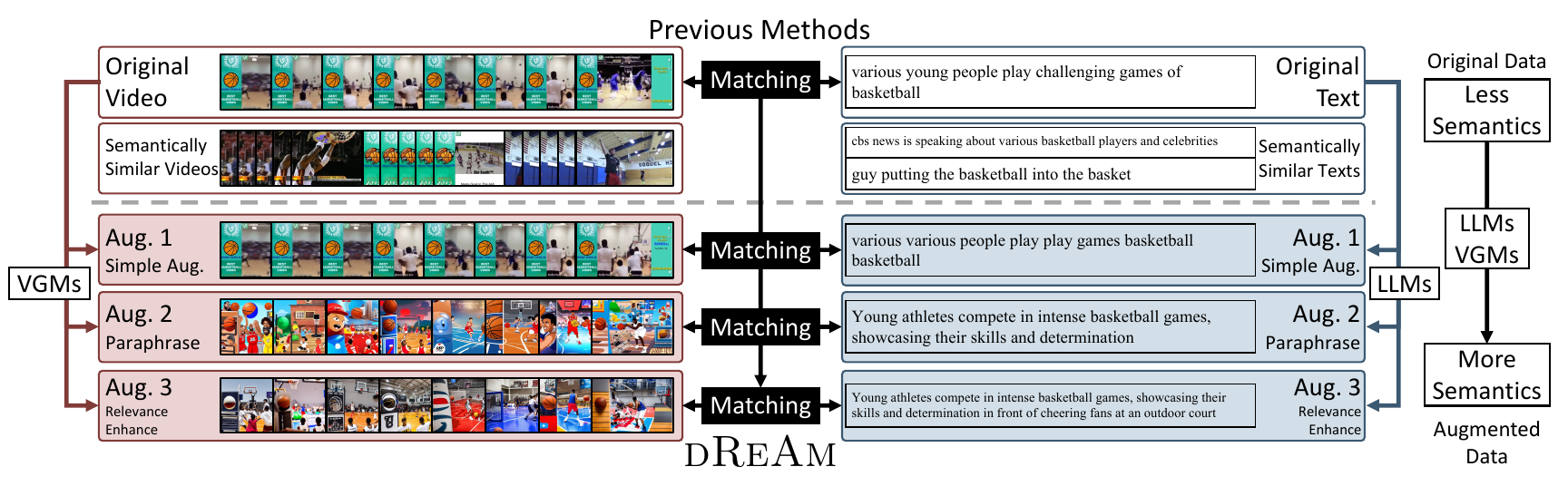}
\caption{
Existing video retrieval works focus on improving representation learning ability by learning from benchmarks that have many semantically similar data points, as shown in the top rows. 
It leads to vague annotations and associations between videos and texts, further hindering the representation learning ability of video-text retrieval models. 
To counteract this issue, in this study, we propose \ours.
Specifically, instead of learning from original noisy data, \ours\ augments data with three proposed augmentation methods, \ie, simple augmentation, augmentation by text paraphrasing and video stylization (``Aug. 2 Parapharse'' in the figure), and augmentation by relevance enhancing (``Aug. 3 Relevance Enhance'' in the figure). 
}\label{fig: framework}
\end{figure*}

\begin{abstract}
    Recent progress in video-text retrieval has been driven largely by advancements in model architectures and training strategies. 
    However, the representation learning capabilities of video-text retrieval models remain constrained by low-quality and limited training data annotations. 
    To address this issue, we present a novel Vi\textsc{\textbf{d}}eo-Text \textsc{\textbf{R}}etrieval Paradigm with R\textsc{\textbf{e}}levance-based \textsc{\textbf{A}}ug\textsc{\textbf{m}}entation, namely \ours, which enhances video and text data using large foundation models to learn more generalized features. 
    Specifically, we first adopt a simple augmentation method, which generates self-similar data by randomly duplicating or dropping subwords and frames. 
    In addition, inspired by the recent advancement in visual and language generative models, we propose a more robust augmentation method through textual paraphrasing and video stylization using large language models (LLMs) and visual generative models (VGMs). 
    To further enrich video and text information, we propose a relevance-based augmentation method, where LLMs and VGMs generate and integrate new relevant information into the original data. 
    Leveraging this enriched data, extensive experiments on several video-text retrieval benchmarks demonstrate the superiority of \ours~over existing methods. 
\end{abstract}

\section{Introduction}
\label{sec:intro}

Video-Text Retrieval (VTR)~\cite{DBLP:journals/ijon/LuoJZCLDL22,DBLP:journals/corr/abs-2111-05610,DBLP:conf/mm/MaXSYZJ22,liu-etal-2022-cross,10.1145/3477495.3531950,DBLP:conf/cvpr/GortiVMGVGY22,fang2022multi,wang-etal-2023-balance,wang-shi-2023-video,yu-etal-2022-ghan} is a fundamental task in visual-language understanding~\cite{9093306,xu-etal-2021-videoclip,park-etal-2022-exposing,miyawaki-etal-2022-scene,fang2023you,fang2023hierarchical,kim2023nice,jian2023invgc}.
The recent progress in VTR is mostly driven by powerful pretraining models~\cite{DBLP:journals/ijon/LuoJZCLDL22,DBLP:journals/corr/abs-2111-05610,DBLP:conf/mm/MaXSYZJ22,liu-etal-2022-cross}, improved retrieval methods~\cite{DBLP:conf/icml/BertasiusWT21,DBLP:conf/cvpr/DongLXJH0W19,DBLP:conf/sigir/JinZZZHZ21}, and the newly emerged large-scale video-language benchmark datasets~\cite{DBLP:conf/cvpr/XuMYR16,chen-dolan-2011-collecting,caba2015activitynet}. 

The most widely adopted VTR paradigm~\cite{DBLP:journals/ijon/LuoJZCLDL22,DBLP:conf/mm/MaXSYZJ22,DBLP:conf/eccv/LiuXXCJ22} learns a joint feature space across the visual and textual modalities, where video and text data are directly compared.
Inspired by the success of CLIP~\cite{radford_learning_2021}, CLIP4Clip~\cite{DBLP:journals/ijon/LuoJZCLDL22} finetunes CLIP~\cite{DBLP:conf/icml/RadfordKHRGASAM21} and investigates three similarity measures for video-sentence contrastive learning, with satisfying retrieval performance. 
Subsequently, X-CLIP~\cite{10.1145/3503161.3547910} introduces a novel multi-grained contrastive learning framework to further enhance the detailed association between video and text modalities. 
Following these pioneering works, many other methods have also been proposed~\cite{wu_cap4video_2023,DBLP:conf/aaai/CaoW0022,liu-etal-2022-cross,DBLP:conf/eccv/LiuXXCJ22,park-etal-2022-normalized,10.1145/3477495.3531950,Fang_2023_ICCV,Wang_2023_ICCV,jin_diffusionret_2023,10.1145/3503161.3547910}. 

Though different modeling or training techniques have been employed to improve the performance on the modeling side, data issues still exist. 
For example, most methods are trained using datasets with one-to-one video-text labels, assuming that the video and text data can be well-aligned one-to-one in the same feature space. 
However, this assumption may not hold tight~\cite{avidan_multi-query_2022} for some popular video-text benchmark datasets~\cite{DBLP:conf/cvpr/XuMYR16,chen-dolan-2011-collecting,caba2015activitynet}, because in real applications, a single video may correspond to multiple valid sentences, and vice versa. 
As shown in \Cref{fig: framework} (an example from MSR-VTT~\cite{DBLP:conf/cvpr/XuMYR16}), the basketball video in \Cref{fig: framework} is paired with the text, ``various young people play challenging games of basketball'', although it can also match with many other semantically similar sentences such as ``guy putting the basketball into the basket'' or ``cbs news is speaking about various basketball players and celebrities''. 
Similarly, one text can also potentially corrrespond to many different videos. 
One possible solution to this mismatch problem is to have better datasets with precise one-to-one video-text pairs. 
However, it is extremely challenging to have a sufficiently large dataset with high-quality due to the nature of the ambiguity of video and text data themselves.

Thus, instead of collecting new high-quality datasets, in this paper, we propose a simple yet effective framework, namely \ours, to enhance the one-to-one matching between video and text by semantically augmenting video and text data. 
As shown in \Cref{fig: framework}, though videos and texts have many semantically similar neighbors, they still differ from each other with minor differences. 
Motivated by the success of data augmentation for better representation learning in computer vision~\cite{DBLP:conf/icml/ChenK0H20} and natural language processing~\cite{gao-etal-2021-simcse}, we utilize data augmentation to enlarge the minor differences between semantically similar data for enhancing the quality of datasets. 

Specifically, we first introduce a simple augmentation method, which generates semantically similar videos and texts through random duplication or deletion of frames or subwords. 
Our experiments show that even such a simple augmentation method can improve the text-to-video Recall@1 on MSR-VTT from 46.1 to 50.8. 
Next, inspired by the success of the latest large foundation models such as large language models (LLMs)~\cite{touvron2023llama,touvron2023llama2,groeneveld2024olmo,brown2020language} and visual generative models (VGMs)~\cite{Saharia2022,zhang2023adding,brooks2022instructpix2pix,wang-etal-2023-t2iat}, we utilize these off-the-shelf models and propose two augmentation strategies, \ie, augmentation by text paraphrasing and video stylization (TPVS) and augmentation by relevance enhancing (RE). 
TPVS employs off-the-shelf large models to generate semantically similar videos and text by stylization (\eg, cartoon style) and text paraphrasing. 
In addition, to infuse video and text with richer information, we introduce a relevance-based augmentation method, where videos and texts are expanded with relevant information given the input video or text. 
Two advanced methods further improve the text-to-video Recall@1 on MSR-VTT from 46.1 to 56.0 and 60.8. 
To the best of our knowledge, we are the first to exploit the latest foundation models to augment data for VTR. 

To understand how our proposed augmentation methods improve VTR performance, extensive experiments on three representative VTR benchmarks show that our proposed \ours\ outperforms our baseline and previous methods by a large margin.

In summary, our contributions are as follows,
\begin{itemize}
    \item We identify the challenge of video-text retrieval as the ambiguous one-to-one labels that hinder learning robust representations. 
    We explore augmentation techniques along several dimensions
    as a way to address this challenge. 
    
    \item Our proposed \ours\ includes three augmentation methods, \ie, simple augmentation (SA), augmentation by text paraphrasing and video stylization (TPVS), and augmentation by relevance enhancing (RE). 
    We are among the pioneers in the use of the latest large language models and visual generative models to assist video-text retrieval.
    
    \item Extensive experiments show that our proposed \ours~ achieves state-of-the-art performances on three popular benchmarks MSR-VTT, MSVD, and ActivityNet. 
\end{itemize}

\section{Related Works}

\textbf{Video-Text Retrieval (VTR).} 
VTR, which involves cross-modal alignment and abstract understanding of temporal images (videos), has been a popular and fundamental task of language-grounding problems~\cite{10.1145/3394171.3413882,DBLP:conf/prcv/WangWXZ20,ijcai2021p156,yu2023multimodal}. 
Inspired by the success of self-supervised pretraining methods~\cite{devlin-etal-2019-bert,radford2019language,DBLP:conf/nips/BrownMRSKDNSSAA20} and vision-language pretraining~\cite{DBLP:conf/eccv/Li0LZHZWH0WCG20,DBLP:conf/nips/Gan0LZ0020,DBLP:conf/cvpr/SinghHGCGRK22} on large-scale unlabeled cross-modal data, recent works~\cite{DBLP:conf/cvpr/LeiLZGBB021,DBLP:journals/corr/abs-2109-04290,DBLP:journals/corr/abs-2111-05610,DBLP:conf/mm/MaXSYZJ22,park-etal-2022-exposing,DBLP:conf/mm/WangXHLJHD22,9878037,10.1145/3477495.3531950,DBLP:conf/cvpr/GortiVMGVGY22} have attempted to pre-train or fine-tune video-text retrieval models in an end-to-end manner. 
Previous methods have focused on improving the representation learning ability by advanced architectures. 
However, due to the nature of benchmarks~\cite{DBLP:conf/cvpr/XuMYR16,chen-dolan-2011-collecting,caba2015activitynet}, learning from such benchmarks makes the learning procedure unstable. 
To address this issue, we propose to augment data using large language models and visual generative models.

\noindent
\textbf{Learning from data augmentation.} 
Data augmentation~\cite{yang2023image}, as an effective way to improve the sufficiency and
diversity of training data, has become a necessary part of the successful application of computer vision~\cite{DBLP:conf/cvpr/MajurskiMPSHJB19,DBLP:conf/cvpr/LiuYLHR23,DBLP:conf/cvpr/ChenL23a,yuan2023semarflow} and natural language processing~\cite{wei-zou-2019-eda,zhou-etal-2022-melm,xu-etal-2021-augnlg,kobayashi-2018-contextual}. 
Similarly, we first introduce a simple augmentation method that randomly duplicates and drops frames and words to generate self-similar data.

\noindent
\textbf{Learning from synthetic data.} 
As the emergence of language and visual generative models~\cite{touvron2023llama,touvron2023llama2,groeneveld2024olmo,brown2020language,Saharia2022,zhang2023adding,brooks2022instructpix2pix,wang-etal-2023-t2iat}, generating data for learning representation has attracted extensive attention recently. 
In natural language processing, large language models have been used for generating data and labels~\cite{you-etal-2023-large,chong-etal-2022-detecting,khalifa-etal-2021-self} for a while. 
It shows an impressive ability to help researchers collect high-quality domain-specific data~\cite{li-etal-2023-two,xiao-etal-2023-freeal}. 
On the other side, the attempt to use visual generative models to train models without any human-annotated data succeeds in image segmentation~\cite{Feng_2023_CVPR}, domain adaptation~\cite{Tang_2023_CVPR,wang2025mitigatingmodalitygapfewshot}, and more~\cite{Zeng_2023_CVPR,Takmaz_2023_ICCV,Cascante-Bonilla_2023_ICCV,Yang_2023_ICCV}. 
Drawing inspiration from these works, we leverage generative models to augment data by caption paraphrasing and video stylization with relevant information conditioned on the original data.

\section{Method}

In this section, we present the definition of video-text retrieval and the details of \ours, along with three proposed simple but effective augmentation methods as shown in \Cref{fig: framework}.

\subsection{Problem Definition}

In this paper, we focus on video-text retrieval (VTR), aiming to learn a pair of encoders that map data from video and text into a common space where they can be directly compared. 
The query and gallery modalities are denoted as $\mathcal{X}$ and $\mathcal{Y}$.  
The (test) gallery, denoted by $G=\{\mathbf{g}_1, \ldots, \mathbf{g}_{N_G}\}$, contains all the embeddings of the gallery data, where $N_G$ is the size of the gallery data. 
In VTR, the gallery data does not overlap with the training data. 
A video is composed of several frames, as $V = [V_{1}, \ldots, V_{N_{frames}}]$, where $N_{frames}$ is the number of frames of that video, and $V_i$ is the $i$-th frame of that video. 
A text is represented by multiple words, as $T = [T_1, \ldots, T_{N_{words}}]$, where $N_{words}$ is the number of (sub-)words, and $T_i$ is the $i$-th (sub-)word.
The goal of VTR is to learn a video encoder $f_{video}(\cdot)$ and a text encoder $f_{text}(\cdot)$ that map video and text into a common space, on which paired video-text data are close.

\subsection{\ours}

The motivation of \ours\ is that low-quality benchmarks~\cite{xu_msr-vtt_2016,chen-dolan-2011-collecting,caba2015activitynet} lead to unsatisfying representation learning.
On the other side, the simple self-augmenting method (SA) improves the retrieval performance, as shown in \Cref{fig: framework,tab: ablation: number of text,tab: ablation: number of video}. 
Inspired by the results, we propose three simple but effective data augmentation methods to enrich data and further boost retrieval performance, as shown in \Cref{fig: framework}. 

Specifically, for each video $V$ and text $T$, we augment it by generating positive views as $\tilde{V}$ and $\tilde{T}$. 
Then, instead of doing multi-query retrieval, we concatenate the positive views with the original data for a fair comparison with previous methods. 
After that, we use a representative VTR method, \ie, X-CLIP~\cite{10.1145/3503161.3547910}, as our learning method for aligning video and text spaces.

\subsubsection{Simple Augmentation (SA)} 
SA targets generating self-similar data without any prior or pretrained models. 
A simple implementation is randomly duplicating and dropping some frames and words without changing the original order. 
Specifically, denoting the original video and text as $V = [V_{1}, \ldots, V_{N_{frames}}]$ and $T = [T_1, \ldots, T_{N_{words}}]$, we sample $\tilde{N}_{frame}$ frames or $\tilde{N}_{words}$ words with replacement to form the augmented videos $\tilde{V}$ and texts $\tilde{T}$. 
For example, for a 2-frame video $V = [V_{1}, V_{2}]$,  we will have three different augmentations, \ie, $[V_{1}, V_{1}]$, $[V_{2}, V_{2}]$, and $[V_{1}, V_{2}]$.
Similarly, for a 2-subword text $T = [T_1, T_2]$, we also have three different augmentations, \ie, $[T_{1}, T_{1}]$, $[T_{2}, T_{2}]$, and $[T_{1}, T_{2}]$.

\subsubsection{Augmentation by Text Paraphrasing and Video Stylization (TPVS)}
The goal of \ours\ is to add as many as possible details in the video and text to enrich the data and thus further improve the representation learning ability for boosting retrieval performance. 
One straightforward approach to enriching data is generating videos and texts based on data from another modality using multi-modal generative models~\cite{DBLP:journals/corr/abs-2206-00169,li_blip_2022,li_blip-2_2023,wang_ofa_2022,reedGenerativeAdversarialText2016,feng-etal-2023-uncovering}.
That could be useful during training, as it brings extrat precise information. 
However, this cannot be used for test time augmentation, as during inference, the data from another modadlity is not available. 
That prompts us to focus on generalizations based on a single modality.

Recently, with the emergence of visual generative models (VGMs) and large language models (LLMs), enriching the video and text data by paraphrasing~\cite{bansal-etal-2023-rethinking} and stylization~\cite{zhang2023adding} become a valid solution. 
Inspired by the advancement in LLMs and VGMs, we propose to generate the paraphrased text and a new style of the original video with standard foundational models.

\paragraph{Text paraphrasing}
Specifically, for augmenting text captions, we use the following prompts:
\lstset{
  basicstyle=\ttfamily,
  columns=fullflexible,
  numbers=none,
  frame=shadowbox,
  breaklines=true,
  breakindent=0pt,
}
\begin{lstlisting}
<@\textcolor{BlueGreen}{This is a hard problem}@>. The following is a caption from a video: ["text"]. Based on this caption, <@\textcolor{Mahogany}{carefully generate a paraphrased caption capturing the key information and main themes in one sentence with up to twenty words}@>:
\end{lstlisting}

Our prompt has two specific designs for generating high-quality paraphrased texts, which are essential for VTR as a more detailed caption will help models learn a precise mapping, as shown below:

\textbf{Interrogative/instructive hints} (\textcolor{BlueGreen}{This is a hard problem}). 
Previous studies~\cite{DBLP:conf/nips/KojimaGRMI22} show that adding short interrogative/instructive sentences to the beginning of a prompt can improve zero-shot performance. 
We add a short sentence, ``this is a hard problem'', at the beginning of our prompt for generating paraphrases and found that this generally improved the quality of paraphrased captions. 

\textbf{Style transfer and contextual captions} (\textcolor{Mahogany}{carefully generate a paraphrased caption capturing the key information and main themes in one sentence with up to twenty words}). 
We add specific guidance on this goal and lead LLMs to generate semantically similar captions without adding too many irrelevant words. 

\paragraph{Video sylization}
For augmenting videos, though recent years have witnessed huge progress in video stylization~\cite{yin-etal-2023-nuwa,Wu_2023_ICCV,Khachatryan_2023_ICCV} and generation~\cite{Gao_2023_CVPR,Ruan_2023_CVPR,Shen_2023_CVPR,Ni_2023_CVPR}, the performance of video generation is still far behind image generation~\cite{Saharia2022,zhang2023adding,brooks2022instructpix2pix,wang-etal-2023-t2iat,Rangwani_2023_CVPR} and stylization~\cite{Kang_2023_CVPR,Yang_2023_CVPR,Liu_2023_CVPR,Li_2023_CVPR,Zhou_2023_CVPR}, due to the high requirement of understanding temporal association and highly informative context. 
Besides, as video generation methods have high computation requirements, due to these issues, instead of employing off-the-shelf video generation methods~\cite{Shen_2023_CVPR,Ni_2023_CVPR,Muaz_2023_ICCV,Wang_2023_ICCV}, we employ image stylization methods~\cite{zhang2023adding,Yang_2023_CVPR}, which show better performance and efficiency. 
Specifically, we use each frame of a video as the input of ControlNet~\cite{zhang2023adding} and generate semantically similar frames without any text guidance using ControlNet under different predefined stylized text prompts. 

\subsubsection{Augmentation by Relevance Enhancing (RE)}  
While TPVS shows satisfying retrieval performance, it restricts the addition of extra information and further hinders the quality of data pairs. 
As LLMs and VGMs show the ability to understand the world, we propose the third augmentation method, Augmentation by Relevance Enhancing (RE), which utilizes ``world models'' to enrich the visual and language information in video-text paired data. 

\paragraph{Text relevance enhancing}
Specifically, for augmenting text captions with the enhanced relevant details, we use the following prompts:
\begin{lstlisting}
<@\textcolor{BlueGreen}{This is a hard problem.}@> The following is a caption from a video: ["text"]. Based on this caption, <@\textcolor{Mahogany}{carefully generate a paraphrased caption capturing the key information and main themes in one sentence with up to twenty words}@> (<@\textcolor{ForestGreen}{feel free to add more relevant details based on your knowledge and specalutaion}@>):
\end{lstlisting}
Compared with the prompt used in TPVS, we have one more special design to incorporate additional information.

\textbf{Encouraging uncertainty} (\textcolor{ForestGreen}{feel free to add more relevant details based on your knowledge and specalutaion}). 
In our prompt design, we aim to encourage the model to include potential uncertainty in the paraphrased texts. 
The uncertainty can be seen as semantically similar information, which is effective for better capturing key features.

\paragraph{Video relevance enhancing}
Similar to TPVS, we employ image stylization methods~\cite{Saharia2022,zhang2023adding,brooks2022instructpix2pix,wang-etal-2023-t2iat}, use each frame of a video as the input of the image stylization model, and generate semantically similar frames as augmented views. 
However, to add relevant visual cues, we use ControlNet~\cite{zhang2023adding} without any text guidance in guess mode. 

\subsection{Base Model and Training Objectives}

\begin{table*}[t!]
\centering
\resizebox{\textwidth}{!}{%
\begin{tabular}{ll|ccccc|ccccc}
\toprule
\multirow{2}{*}{Methods}  & \multirow{2}{*}{Venue}          & \multicolumn{5}{c|}{Text-to-Video Retrieval}                                        & \multicolumn{5}{c}{Video-to-Text Retrieval}                                        \\
        &     & R@1$\uparrow$ & R@5$\uparrow$ & R@10$\uparrow$ & MdR$\downarrow$ & MnR$\downarrow$ & R@1$\uparrow$ & R@5$\uparrow$ & R@10$\uparrow$ & MdR$\downarrow$ & MnR$\downarrow$ \\
\midrule
VLM~\cite{xu-etal-2021-vlm} & ACL'21 & 28.1 & 55.5 & 67.4 & 4.0 & - & -&-&-&-&-\\
VideoCLIP~\cite{xu-etal-2021-videoclip} &EMNLP'21 &  30.9 & 55.4 & 66.8 & - & -&-&-&-&- & -\\
LGDN~\cite{DBLP:conf/nips/LuDFH022} & NeurIPS'22 & 43.7 & 71.4 & 80.3 & 2.0 & - & 42.6 & 71.6 & 80.6 & 2.0 & - \\
\midrule
\rowcolor{gray!10}\multicolumn{12}{l}{\textit{BLIP-based}} \\ 
BLIP~\cite{li_blip_2022} & ICML'22 & 41.4 & 63.3 & 72.8 & 2.0 & - & -&-&-&-&- \\
LiteVL-S~\cite{chen-etal-2022-litevl} & EMNLP'22 & 46.7 & 71.8 & 81.7 & 2.0 & - & -&-&-&-&-\\
LiteVL-L~\cite{chen-etal-2022-litevl} & EMNLP'22 & 50.8 & 76.3 & 84.4 & 2.0 & - & -&-&-&-&-\\
\midrule
\rowcolor{gray!10}\multicolumn{12}{l}{\textit{ViT (CLIP)-based}}\\
CLIP~\cite{radford_learning_2021}               &    ICML'21        & 31.2          & 53.7          & 64.2           & 4.0               & -               & 27.2          & 51.7          & 62.6           & 5.0               & -               \\
CLIP4Clip~\cite{DBLP:journals/ijon/LuoJZCLDL22} & NeurComp'22 & 44.5          & 71.4          & 81.6           & 2.0               & 15.3            & -             & -             & -              & -               & -               \\
VCM~\cite{DBLP:conf/aaai/CaoW0022}         & AAAI'22        & 43.8          & 71.0          & -           & 2.0             & 14.3            & 45.1          & 72.3          & 82.3           & 2.0             & 10.7            \\
DiscreteCodebook~\cite{liu-etal-2022-cross}& ACL'22   & 43.4          & {72.3}          & 81.2           & -               & {14.8}            & 42.5          & 71.2          & 81.1           & -               & 12.0            \\
X-Pool~\cite{DBLP:conf/cvpr/GortiVMGVGY22}    & CVPR'22 & 46.9          & 72.8          & 82.2           & 2.0               & 14.3            & -             & -             & -              & -               & -               \\
TS2-Net~\cite{DBLP:conf/eccv/LiuXXCJ22}  & ECCV'22  & {47.0}          & {74.5}          & {83.8}           & 2.0             & {13.0}            & 45.3          & {74.1}          & {83.7}           & 2.0             & {9.2}             \\
NCL~\cite{park-etal-2022-normalized}    & EMNLP'22 & 43.9 & 71.2 & 81.5 & 2.0 & 15.5 & 44.9 & 71.8 & 80.7 & 2.0 & 12.8      \\
Align\&Tell~\cite{9878037}  & TMM'22       & 45.2          & 73.0          & 82.9           & 2.0             & -               & 43.4          & 70.9          & 81.8           & 2.0             & -               \\
TABLE~\cite{chen_tagging_2023} & AAAI'23 & 47.1 & 74.3 & 82.9 & 2.0 & 13.4 & 47.2 & {74.2} & {84.2} & 2.0 & 11.0\\
VOP~\cite{huang_vop_2023} & CVPR'23 & 44.6 & 69.9 & 80.3 & 2.0 & 16.3 & 44.5 & 70.7 & 80.6 & 2.0 & 11.5 \\
PIDRo~\cite{guan_pidro_2023} & CVPR'23 & 48.2 & 74.9 & 83.3 & 2.0 & 12.6 & 47.4 & 74.8 & 84.1 & 2.0 & 8.7 \\
HBI~\cite{Jin_2023_CVPR} & CVPR'23 & 48.6 & 74.6 & 83.4 & 2.0 & 12.0 & 46.8 & 74.3 & 84.3 & 2.0 & 8.9 \\
UATVR~\cite{Fang_2023_ICCV} & CVPR'23 & 47.5 & 73.9 & 83.5 & 2.0 & 12.3 & 46.0 & 73.7 & 82.8 & 2.0 & 8.7 \\
Cap4Video~\cite{wu_cap4video_2023} & ICCV'23 & 49.3 & 74.3 & 83.8 & 2.0 & 12.0 & 47.1 & 73.7 & 84.3 & 2.0 & 8.7 \\
UCoFiA~\cite{Wang_2023_ICCV} & ICCV'23 & 49.4 & 72.1 & - & -  & 12.9 & 47.1 & 74.3 & - & - & - \\
ProST~\cite{li_progressive_2023} & ICCV'23 & 48.2 & 74.6 & 83.4 & 2.0 & 12.4 & 46.3 & 74.2 & 83.2 & 2.0 & 8.7 \\
DiffusionRet~\cite{jin_diffusionret_2023} & ICCV'23 & 49.0 & 75.2 & 82.7 & 2.0 & 12.1 & 47.7 & 73.8 & 84.5 & 2.0 & 8.8\\
RAP~\cite{cao-etal-2024-rap} & ACL'24 & 44.8 & 71.4 & 81.5 & - & 14.4 & 44.0 & 71.9 & 82.4 & - & 10.1 \\
T-MASS~\cite{10655719} & CVPR'24 & 50.2 & 75.3 & 85.1 & 1.0  & 11.9 & - &  - &  - &  - &  - \\
\midrule
X-CLIP~\cite{10.1145/3503161.3547910} (Baseline)  & ACM MM'22  & 46.1          & {74.3}          & {83.1}           & 2.0             & {13.2}            & {46.8}          & 73.3          & {84.0}           & 2.0             & {9.1}             \\
\rowcolor{green!10}\ours  &    & \textbf{60.8}   & \textbf{84.5}   & \textbf{91.4}   & \textbf{1.0}    & \textbf{5.8}    & \textbf{60.6}   & \textbf{85.2}   & \textbf{92.5}   & \textbf{1.0}    & \textbf{5.9}             \\
\bottomrule
\end{tabular}%
}
\caption{
Video-Text retrieval results on MSR-VTT. 
The best results are marked in \textbf{bold}. 
``NeurComp'' refers to Neurocomputing. 
}
\label{tab:msr-vtt}
\end{table*}

In this part, we present a general VTR framework widely used by previous methods~\cite{DBLP:journals/ijon/LuoJZCLDL22,liu-etal-2022-cross}. 
With this paradigm, we obtain two representations for video and text modalities, \ie, video representation $\mathbf{e}_{v}$ and text representation $\mathbf{e}_{t}$ by modality-dependent encoders $f_{video}(\cdot)$ and $f_{text}(\cdot)$. 
Then, the similarity between the video and the text $\operatorname{sim}(\mathbf{e}_{v}, \mathbf{e}_{t})$ is calculated by the cosine similarity $s = cosine(\mathbf{e}_{v}, \mathbf{e}_{t})$.
Finally, the retrieved data is ranked based on the cosine similarity to the query input. 

The training objective is the contrastive loss.
Following Clip4Clip~\cite{DBLP:journals/ijon/LuoJZCLDL22}, we employ the symmetric InfoNCE loss
as,
\begin{align*}
    \ell_{sim} = & \ell_{v2t} + \ell_{t2v} \\
    = &- \frac{1}{N} \sum_{i \in [N]} \log \frac{\exp(s_{i,i})}{\sum_{j\in[N]} \exp(s_{i,j})} \\
    &- \frac{1}{N} \sum_{i \in [N]} \log \frac{\exp(s_{i,i})}{\sum_{j\in[N]} \exp(s_{j,i})}\,,
\end{align*}
where $s_{i,j}$ is similarity between $i$-th video and $j$-th text and $N$ is the number of paired data.

\section{Experiments}

\noindent
\textbf{Benchmarks.} 
To evaluate the proposed \ours, we use three representative VTR benchmarks, \ie, MSR-VTT~\cite{DBLP:conf/cvpr/XuMYR16}, MSVD~\cite{chen-dolan-2011-collecting}, and ActivityNet~\cite{caba2015activitynet}. 
Details are deferred to the Appendix due to the limitation of space.

\noindent
\textbf{Evaluation Protocols. }
To evaluate the retrieval performance of our proposed \ours, we use recall at Rank K (R@K, higher is better), median rank (MdR, lower is better), and mean rank (MnR, lower is better) as retrieval metrics, which are widely used in previous retrieval works~\cite{DBLP:conf/icml/RadfordKHRGASAM21,DBLP:journals/ijon/LuoJZCLDL22,DBLP:conf/mm/MaXSYZJ22}. 

\noindent
\textbf{Implementation Details. }
Our baseline (base model) is X-CLIP~\cite{DBLP:conf/mm/MaXSYZJ22}. 
Following \citet{DBLP:journals/ijon/LuoJZCLDL22,DBLP:conf/mm/MaXSYZJ22}, we use a standard vision transformer~\cite{DBLP:conf/iclr/DosovitskiyB0WZ21} with $12$ layers that are initialized with the public CLIP~\cite{DBLP:conf/icml/RadfordKHRGASAM21} checkpoints. 
We use SeqTransformer as the temporal encoder, similar to \cite{DBLP:journals/ijon/LuoJZCLDL22}.
We directly use the text encoder of CLIP as our text encoder, which is also initialized with the public CLIP checkpoints. 
All models are optimized for 5 epochs on MSR-VTT and MSVD, and for ActivityNet, the models are trained for 20 epochs. 
We use AdamW~\cite{DBLP:conf/iclr/LoshchilovH19} with a weight decay of 0.2 and decay the learning rate using a cosine schedule~\cite{DBLP:conf/iclr/LoshchilovH17}, following the method used in CLIP~\cite{DBLP:conf/icml/RadfordKHRGASAM21}. 
For all experiments, we uniformly sample 12 frames from every video, resizing each frame to 224x224 as per previous works~\cite{DBLP:journals/ijon/LuoJZCLDL22,DBLP:conf/mm/MaXSYZJ22}. 
For text augmentation, we use LLaMA2~\cite{touvron2023llama2}, while for video frame augmentation, we employ ControlNet~\cite{zhang2023adding}.

\subsection{Quantitative Results}
In this part, we present a series of experiments on MSR-VTT, MSVD, and ActivityNet to demonstrate the effectiveness of \ours\ in \Cref{tab:msr-vtt,tab: msvd,tab: activitynet,tab: activitynet full,tab: msvd full}.

\smallskip \noindent 
\textbf{MSR-VTT.} 
The results are shown in \Cref{tab:msr-vtt}. 
\ours\ significantly outperforms all previous methods across different retrieval metrics, achieving remarkable top scores with a Recall@1 of 60.8 and 60.6, Recall@5 of 84.5 and 85.2, and Recall@10 of 91.4 and 92.5, for text-to-video and video-to-text, respectively. 
This leap in performance highlights the effectiveness of \ours, setting a new benchmark for the field. 

\smallskip \noindent 
\textbf{MSVD.}
Corresponding results are shown in \Cref{tab: msvd,tab: msvd full}. 
With a Text-to-Video Retrieval Recall@1 of 61.6, Recall@5 of 87.1, and Recall@10 of 93.2, alongside a MdR and MnR of 1.0 and 5.6 respectively, \ours\ establishes new SOTAs. 

\smallskip \noindent 
\textbf{ActivityNet.}
Corresponding results are shown in \Cref{tab: activitynet,tab: activitynet full}. 
\ours\ achieves the highest scores across both text-to-video and video-to-text retrieval tasks, with Text-to-Video Retrieval scores of Recall@1 at 59.1.

\begin{table}[t!]
\centering
\resizebox{\columnwidth}{!}{%
\begin{tabular}{ll|ccccc}
\toprule
                    \multirow{2}{*}{Methods}  & \multirow{2}{*}{Venue}& \multicolumn{5}{c}{Text-to-Video Retrieval}            \\
                    &           & R@1$\uparrow$ & R@5$\uparrow$ & R@10$\uparrow$ & MdR$\downarrow$ & MnR$\downarrow$  \\
                               \midrule
CLIP4Clip & NeurComp'22 & 45.2          & 75.5          & 84.3     &   2.0 & 10.3       \\
CLIP2Video    &    Arxiv'21             & 47.0          & 76.8          & 85.9    & 2.0      & 9.6     \\
X-Pool       & CVPR'22                  & {47.2}          & 77.4          & {86.0}     &  -    & {9.3}     \\
NCL    & EMNLP'22 & 47.8 & 77.5 & 85.9 & 2.0 & 9.9 \\
CenterCLIP  & SIGIR'22  & 47.6 & 76.8 & 85.6 & 2.0 & 9.9   \\
TABLE & AAAI'23 & 49.9 & 79.3 & 87.4 & 2.0 & 9.1 \\
PIDRo & CVPR'23 & 47.5 & 77.5 & 86.0 & 2.0 & 9.2   \\
UATVR & CVPR'23 & 46.0 & 76.3 & 85.1 & 2.0 & 10.4   \\
Cap4Video & ICCV'23 & 51.8 & 80.8 & 88.3 & 1.0 & 8.3   \\
UCoFiA & ICCV'23 & 47.4 & 77.6 & - & -  & 9.6  \\
DiffusionRet & ICCV'23 & 46.6 & 75.9 & 84.1 & 2.0 & 15.7\\
RAP & ACL'24 & 49.8 & 78.2 & 86.1 & -& 9.7 \\
\midrule
X-CLIP    & ACM MM'22  & 47.1          & {77.8}          & -     &  - & {9.5}   \\
\rowcolor{green!10}\ours (ViT-B/32) & &   \textbf{61.6} & \textbf{87.1} & \textbf{93.2} & \textbf{1.0} & \textbf{5.6}   \\
\bottomrule
\end{tabular}%
}
\caption{Text-to-Video retrieval results on MSVD. 
Best in \textbf{bold}. 
}
\label{tab: msvd}
\end{table}

\begin{table}[t!]
\centering
\resizebox{\columnwidth}{!}{%
\begin{tabular}{ll|ccccc}
\toprule
                    \multirow{2}{*}{Methods}  & \multirow{2}{*}{Venue}& \multicolumn{5}{c}{Text-to-Video Retrieval}              \\
                    &           & R@1$\uparrow$ & R@5$\uparrow$ & R@10$\uparrow$ & MdR$\downarrow$ & MnR$\downarrow$ \\
                               \midrule
CLIP4Clip & NeurComp'22 & 40.5          & 72.4          & 98.1           &   2.0 & 7.4 \\
VCM        & AAAI'22                    & 40.8          & 72.8          & -        &    2.0  & {7.3}       \\
TS2-Net  & ECCV'22                      & 41.0          & 73.6          & {84.5}        & 2.0  & 8.4    \\
NCL    & EMNLP'22 & 45.9 & 76.8 & 98.3 & 2.0 & 6.7 \\
Align\&Tell   & TMM     & 42.6          & 73.8          & -         &   2.0  & -      \\
CenterCLIP  & SIGIR'22  & 43.5 & 75.0 & 85.9 & 2.0 & 6.9   \\
PIDRo & CVPR'23 & 44.9 & 74.5 & 86.1 & 2.0 & 6.4 \\
HBI & CVPR'23 & 42.2 & 73.0 & 84.6 & 2.0 & 6.6  \\
UCoFiA & ICCV'23 & 45.7 & 76.0 & - & -  & 6.6 \\
DiffusionRet & ICCV'23 & 45.8 & 75.6 & 86.3 & 2.0 & 6.5 \\
RAP & ACL'24 & 48.4 & 76.2 & 86.4 & - &  7.0 \\
\midrule
X-CLIP    & ACM MM'22          & {44.3}          & {74.1}          & -    &     -     & 7.9        \\
\rowcolor{green!10}\ours   &  &   \textbf{59.1}    &   \textbf{82.9} & \textbf{95.3} & \textbf{1.0} & \textbf{5.2}   \\
\bottomrule
\end{tabular}%
}
\caption{Text-to-Video retrieval results on ActivityNet. Best in \textbf{bold}. 
}
\label{tab: activitynet}
\end{table}

\begin{figure}[t!]
\centering
\includegraphics[width=1\columnwidth]{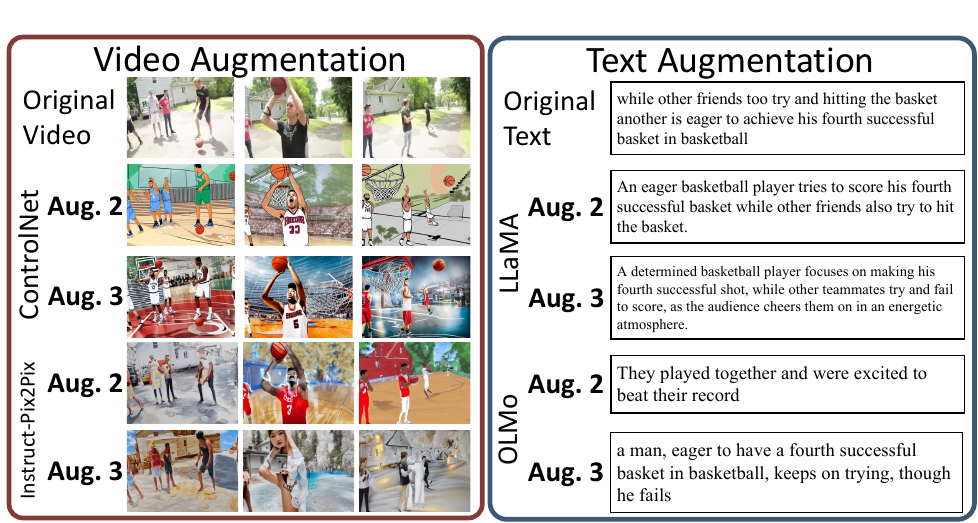}
\caption{
Qualitative examples of data generated by \ours. 
``Aug. 2'' and ``Aug. 3'' refer to augmentation by text paraphrasing and video stylization and augmentation by relevance enhancing. 
}\label{fig: qualitative}
\end{figure}

\begin{figure}[h!]
\centering
\includegraphics[width=1\columnwidth]{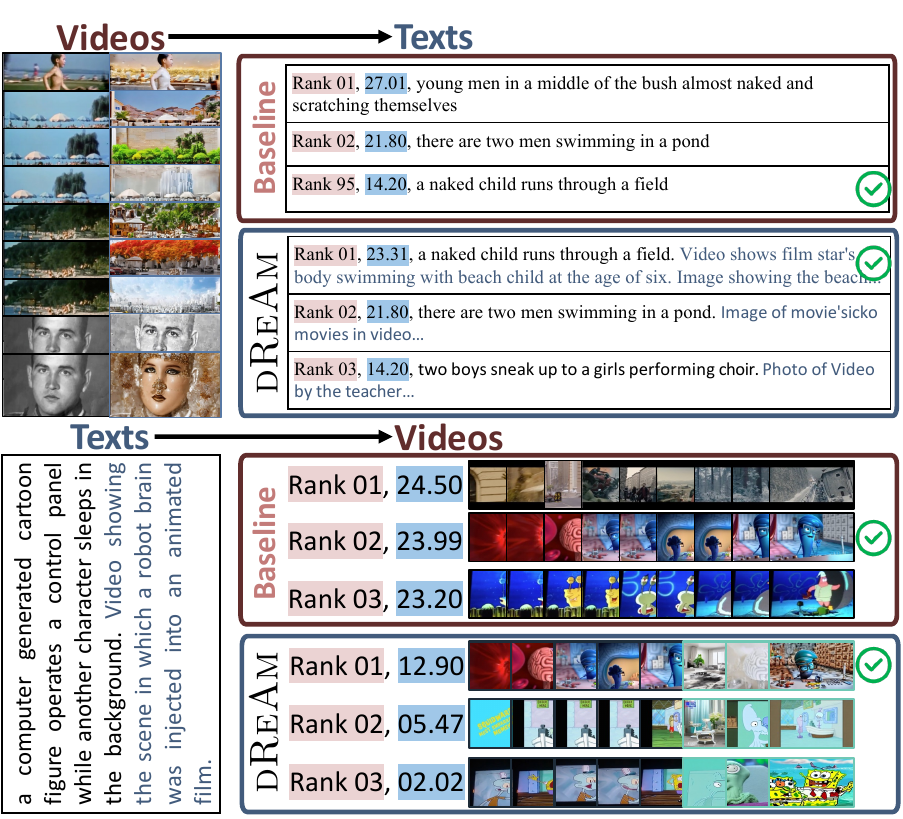}
\caption{
Retrieval examples by the baseline and \ours. 
``Rank x'' means that the example is ranked at $x$. 
The numbers in blue represent the similarity to the query.
Texts in blue and video frames surrounded by green lines are augmented data. 
}\label{fig: retrieval examples}
\end{figure}

\subsection{Qualitative Results}

\textbf{Quality of augmented data}. 
To qualitatively validate the effectiveness of \ours, we present examples of augmented data in \Cref{fig: qualitative,fig: qualitative full}, respectively. 
For paraphrasing text, we employ LLaMA2~\cite{touvron2023llama2} and OLMo~\cite{groeneveld2024olmo}. 
For generating semantically similar video frames, we use ControlNet~\cite{zhang2023adding} and Instruct-Pix2Pix~\cite{brooks2022instructpix2pix}.
We notice that for text paraphrasing, both LLaMA and OLMo are able to grasp the main idea based on the input and generate semantically similar texts. 
For video frame generation, though the generation for a frame is independent of other frames, we still observe that ControlNet can generate frames within a similar style, while the frames generated by Instruct-Pix2Pix are always in different styles. 
A failure case is the second frame in the last row where a man is holding a human head as a basketball. 

\noindent
\textbf{Retrieval examples.}
To qualitatively validate the effectiveness of \ours, we present examples of video-to-text and text-to-video retrieval on MSR-VTT in \Cref{fig: retrieval examples}. 
The retrieval results show the satisfactory performance of \ours, benefiting from the augmented semantics, compared with the baseline. 
While the baseline struggles with matching, \ours\ demonstrates precise identification of objects (\textit{computer}) and humans (\textit{child}), indicating its proficiency in capturing intricate details. 

\subsection{Ablation Studies}

In this section, we present the ablation studies on \ours\ regarding the number of paraphrased texts and generated videos on MSR-VTT utilizing \ours\ with X-CLIP (ViT-B/32) as the base model. 
Due to the space limitation, the results on video-to-text are presented in \Cref{tab: ablation: number of text full,tab: appendix: different image gen,tab: appendix: different llms full,tab: ablation: number of video full}.

\noindent
\textbf{Number of paraphrased texts.} 
\Cref{tab: ablation: number of text} offers an insightful look into the effects of varying the number of augmented texts. 
With simple augmentation, a gradual increase in performance is seen with the number of texts. 
Moving to augmentation by text paraphrasing and video stylization, the performance leaps further, highlighting the value of leveraging external, sophisticated models to enrich the dataset. 
The last one, augmentation by relevance enhancing, showcases the most significant performance boosts, especially with 5 texts, achieving the highest Recall@1 of 60.8, along with the best Recall@5, Recall@10, and the lowest MdR and MnR scores.

\begin{table}[t!]
\centering
\resizebox{\columnwidth}{!}{%
\begin{tabular}{c|ccccc}
\toprule
\multirow{2}{*}{\# of Text}  & \multicolumn{5}{c}{Text-to-Video Retrieval}\\
    & R@1$\uparrow$ & R@5$\uparrow$ & R@10$\uparrow$ & MdR$\downarrow$ & MnR$\downarrow$\\\midrule
Baseline &    46.1          & {74.3}          & {83.1}           & 2.0             & {13.2}       \\
\midrule
\rowcolor{gray!10}\multicolumn{6}{c}{\textit{Augmentation 1: Simple Augmentation}}\\
1 & 50.8 & 76.7 & 84.6 & 1.0 & 11.1  \\
2 & 50.1 & 75.3 & 84.1 & 1.0 & 12.2 \\
3 & 51.3 & \textbf{77.0} & \textbf{85.5} & 1.0 & 11.3 \\
4 & 51.6 & 76.4 & 84.7 & 1.0 & 11.5 \\
5 & \textbf{52.0} & 76.5 & 84.7 & 1.0 & \textbf{11.0} \\
\midrule
\rowcolor{gray!10}\multicolumn{6}{c}{\textit{Augmentation 2: Text Paraphrasing and Video Stylization}}\\
1 & \textbf{56.1} & \textbf{81.3} & \textbf{89.4} & 1.0 & 9.7 \\
2 & 55.7 & 80.8 & 88.8 & 1.0 & \textbf{9.6}\\
3 & 56.0 & 80.5 & 88.7 & 1.0 & 10.0\\
\midrule
\rowcolor{gray!10}\multicolumn{6}{c}{\textit{Augmentation 3: Relevance Enhaching}}\\
1 & 56.7 & 83.0 & 90.1 & 1.0 & 8.1 \\
2 & 60.1 & 83.1 & 90.1 & 1.0 & {5.9}\\
3 & {60.4} & 85.3 & 91.0 & 1.0 & \textbf{5.8}\\
4 & 59.2 & 84.0 & {91.1} & 1.0 & 7.1 \\
\rowcolor{green!10}5 & \textbf{60.8}   & \textbf{84.5}   & \textbf{91.4}   & 1.0    & \textbf{5.8} \\
\bottomrule
\end{tabular}%
    }
\caption{
Text-to-video retrieval performance with different numbers of augmented captions using three augmentation methods on MSR-VTT. 
Best in \textbf{bold}.
}
\label{tab: ablation: number of text}
\end{table}

\begin{table}[t!]
\centering
\resizebox{\columnwidth}{!}{%
\begin{tabular}{c|ccccc}
\toprule
\multirow{2}{*}{\# of Video}  & \multicolumn{5}{c}{Text-to-Video Retrieval}                                   \\
    & R@1$\uparrow$ & R@5$\uparrow$ & R@10$\uparrow$ & MdR$\downarrow$ & MnR$\downarrow$\\\midrule
Baseline &    46.1          & {74.3}          & {83.1}           & 2.0             & {13.2}   \\
\midrule
\rowcolor{gray!10}\multicolumn{6}{c}{\textit{Augmentation 1: Simple Augmention}}\\
1 & \textbf{50.4} & 75.8 & 84.9 & 1.0 & 11.2 \\
2 & 48.7 & 75.6 & 84.3 & 2.0 & 11.4 \\
3 & 50.0 & 75.0 & 84.2 & 1.5 & 12.4\\
4 & 49.8 & 74.4 & 82.8 & 2.0 & 11.6\\
5 & \textbf{50.4} & \textbf{77.2} & \textbf{86.0} & 1.0 & \textbf{10.7} \\
\midrule
\rowcolor{gray!10}\multicolumn{6}{c}{\textit{Augmentation 2: Text Paraphrasing and Video Stylization}}\\
1 & 53.8 & \textbf{80.7} & 88.6 & 1.0 & 9.9  \\
2 & 54.4 & 79.3 & 87.4 & 1.0 & 10.1 \\
3 & \textbf{54.7} & 80.3 & \textbf{88.9} & 1.0 & \textbf{9.3}  \\
\midrule
\rowcolor{gray!10}\multicolumn{6}{c}{\textit{Augmentation 3: Relevance Enhaching}}\\
1 & 60.4 & 84.4 & \textbf{91.4} & 1.0 & 6.2 \\
2 & 60.0 & 84.0 & 91.2 & 1.0 & 6.4  \\
\rowcolor{green!10}3 & \textbf{60.8}   & \textbf{84.5}   & \textbf{91.4}   & 1.0    & \textbf{5.8}   \\
\bottomrule
\end{tabular}
}
\caption{
Text-to-video retrieval performance with different numbers of generated videos using three augmentation methods on MSR-VTT. 
Best in \textbf{bold}. 
}
\label{tab: ablation: number of video}
\end{table}

\noindent
\textbf{Number of generated videos.} 
\Cref{tab: ablation: number of video} presents an ablation study on the impact of different numbers of generated videos. 
For simple augmentation, we notice considerable improvements, particularly with 5 videos, achieving an R@1 of 50.4 on text-to-video. 
When moving to augmentation by text paraphrasing and video stylization, it further elevates the performance, with the best results observed when using 3 videos, where T2V R@1 reaches 54.7 and V2T R@1 peaks at 56.4. 
This suggests that leveraging foundation models for augmentation can significantly impact retrieval effectiveness, likely due to the richer, more diverse semantic representations introduced.
The last strategy, augmentation by relevance enhancing, achieves the highest T2V R@1 of 60.8, alongside the best R@5 and R@10 scores, underscoring the efficacy of augmentation in capturing diverse semantic content.

\begin{table}[t!]
\centering
\resizebox{\columnwidth}{!}{%
\begin{tabular}{lc|ccccc}
\toprule
 \multirow{2}{*}{Methods}  & \multirow{2}{*}{LLMs}          & \multicolumn{5}{c}{Text-to-Video Retrieval}                                          \\
        &     & R@1$\uparrow$ & R@5$\uparrow$ & R@10$\uparrow$ & MdR$\downarrow$ & MnR$\downarrow$ \\
\midrule
Baseline   & -  & 46.1          & {74.3}          & {83.1}           & 2.0             & {13.2}               \\
\midrule
\rowcolor{gray!10}\multicolumn{7}{c}{\textit{Augmentation 2: Augmentation by Text Paraphrasing and Video Stylization}}\\
 \multirow{2}{*}{\ours}  & LLaMA-2-7b-chat-hf & 54.7 & 80.3 & 88.9 & 1.0 & 9.3\\
   &  OLMo-7b &     53.2 &  77.9 & 86.3 & 1.0 & 9.8 \\
\midrule
\rowcolor{gray!10}\multicolumn{7}{c}{\textit{Augmentation 3: Augmentation by Relevance Enhaching}}\\
\rowcolor{green!10} \cellcolor{white!10}{\multirow{2}{*}{\ours}}  &   LLaMA-2-7b-chat-hf & \textbf{60.8}   & \textbf{84.5}   & \textbf{91.4}   & \textbf{1.0}    & \textbf{5.8}      \\
   &  OLMo-7b &       58.7 & 81.2 & 89.5 & 1.0 & 7.5   \\
\bottomrule
\toprule
 \multirow{2}{*}{Methods}  & \multirow{2}{*}{VGMs}          & \multicolumn{5}{c}{Text-to-Video Retrieval}   \\
        &     & R@1$\uparrow$ & R@5$\uparrow$ & R@10$\uparrow$ & MdR$\downarrow$ & MnR$\downarrow$\\
\midrule
Baseline   & -  & 46.1          & {74.3}          & {83.1}           & 2.0             & {13.2}         \\
\midrule
\rowcolor{gray!10}\multicolumn{7}{c}{\textit{Augmentation 2: Augmentation by Text Paraphrasing and Video Stylization}}\\
 \multirow{2}{*}{\ours} & ControlNet  &  54.7 & 80.3 & 88.9 & 1.0 & 9.3\\
   &  instruct-pix2pix &     56.7 & 83.0 & 90.1 & 1.0 & 8.1       \\
\midrule
\rowcolor{gray!10}\multicolumn{7}{c}{\textit{Augmentation 3: Augmentation by Relevance Enhaching}}\\
\rowcolor{green!10} \cellcolor{white!10}{\multirow{2}{*}{\ours}}  &   ControlNet & \textbf{60.8}   & \textbf{84.5}   & \textbf{91.4}   & \textbf{1.0}    & \textbf{5.8}      \\
   &  instruct-pix2pix &      57.8 & 82.9 & 90.2 & 1.0 & 7.9    \\
\bottomrule
\end{tabular}%
}
\caption{
Text-to-video retrieval results on MSR-VTT using different image generation methods for generating stylized video frames. 
Best in \textbf{bold}. 
}
\label{tab: appendix: different text and image gen}
\end{table}

\begin{table}[t!]
\centering
\resizebox{\columnwidth}{!}{%
\begin{tabular}{c|ccc|ccc}
\toprule
  & \multicolumn{3}{c|}{Text-to-Video Retrieval}         &\multicolumn{3}{c}{Video-to-Text Retrieval}                                   \\
    & R@1$\uparrow$ & R@5$\uparrow$  & MnR$\downarrow$ & R@1$\uparrow$ & R@5$\uparrow$  & MnR$\downarrow$\\\midrule
UCoFiA & 49.4	&72.1	&12.9	&47.1	&74.3	&11.4 \\
\rowcolor{green!10}\ours (UCOFIA) & \textbf{58.6}	&\textbf{84.3}&	\textbf{6.1}&	\textbf{58.0}	&\textbf{83.6}	&\textbf{5.1} \\
\bottomrule
\end{tabular}
}
\caption{
Text-to-video retrieval performance with UCoFiA as the base model on MSR-VTT. 
Best in \textbf{bold}. 
}
\label{tab: ablation: ucofia}
\end{table}

\noindent\textbf{Choice of LLM for paraphrasing.} 
To understand how LLMs impact retrieval performance, we also use OLMo~\cite{groeneveld2024olmo} for generating paraphrased captions, as shown in \Cref{tab: appendix: different text and image gen,fig: qualitative}. 
The data showcases a notable improvement when using LLMs over the baseline, with LLaMA achieving the highest performance across all metrics compared to OLMo, underscoring its superiority in understanding and generating nuanced paraphrased captions that significantly benefit retrieval accuracy.

\noindent \textbf{Choice of image generation methods.} 
We also use Instruct-Pix2Pix~\cite{brooks2022instructpix2pix} for generating video frames, as shown in \Cref{tab: appendix: different text and image gen,fig: qualitative}. 
It underscores the superiority of the ControlNet in the Augmentation by relevance enhancing, marking it with 60.8 for Recall@1.

\noindent \textbf{Generalization on more base models.} 
As shown in \cref{tab: ablation: ucofia}, we also employ UCoFiA as our base model. 
Results show that \ours shows strong generalization ability as the performance of UCOFIA is improved by a large margin.

\section{Conclusion}
In this paper, we proposed a novel video-text learning paradigm, \ours, which effectively aligned video and text spaces using generated information conditioned on original data. 
First, we showed a simple but effective method, self-augmenting, which generated self-similar data without any parameters by randomly duplicating or removing frames and subwords, significantly enhancing representation learning and mitigating overfitting issues commonly observed with current models.
Second, inspired by the advancement in large language models (LLMs) and video generative models (VGMs), \ours\ employed a novel augmentation method, which augmented data through paraphrasing captions and transferring video styles. 
Last, to enrich video and text data with relevant information, we proposed to augment data with relevance, which encouraged LLMs and VGMs to inject relevant information, and further novel information into the generated data. 
This method significantly enriched the data pool, contributing to the robustness and depth of the learned representations. 
Finally, comprehensive experiments conducted on several video-text retrieval benchmarks underline the superior performance of \ours.

\section*{Limitations}
It would be interesting to test whether the proposed augmentation methods can improve the performance of more base models. 
Moreover, limited by computation resources, we only use image-generation methods to augment videos. 
It would be interesting to investigate the power of video generation methods that consider temporal association in the input. 
Inspired by the recent progress of vision-language models (VLMs), such as BLIP, InternVL, and LLaVA, we present the preliminary results using those powerful VLMs. 
It is promising to employ powerful VLMs for video retrieval with advance augmentation techniques.

\section*{Ethical Considerations}
As visual generative models (VGMs) and large language models (LLMs) are used in this study to provide data augmentations, the bias of VGMs and LLMs could be attributed to the bias of retrieval methods. 
On the other side, the proposed methods do not have any potential risks.

\bibliography{anthology,custom,references,new}
\newpage\clearpage
\appendix

\appendix

In this technical appendix, we present the experimental results on Video-to-Text retrieval and detailed qualitative results.

\section{Experiments}
\subsection{Details of Benchmark Datasets}

\textbf{MSR-VTT}~\cite{DBLP:conf/cvpr/XuMYR16} contains 10,000 videos with length varying from 10 to 32
seconds, each paired with about 20 human-labeled captions. 
Following the evaluation protocol from previous works~\cite{DBLP:conf/eccv/YuKK18,DBLP:conf/iccv/MiechZATLS19}, we use the training-9k / test 1k-A splits for training and testing, respectively. 

\begin{figure*}[h!]
\centering
\includegraphics[width=\textwidth]{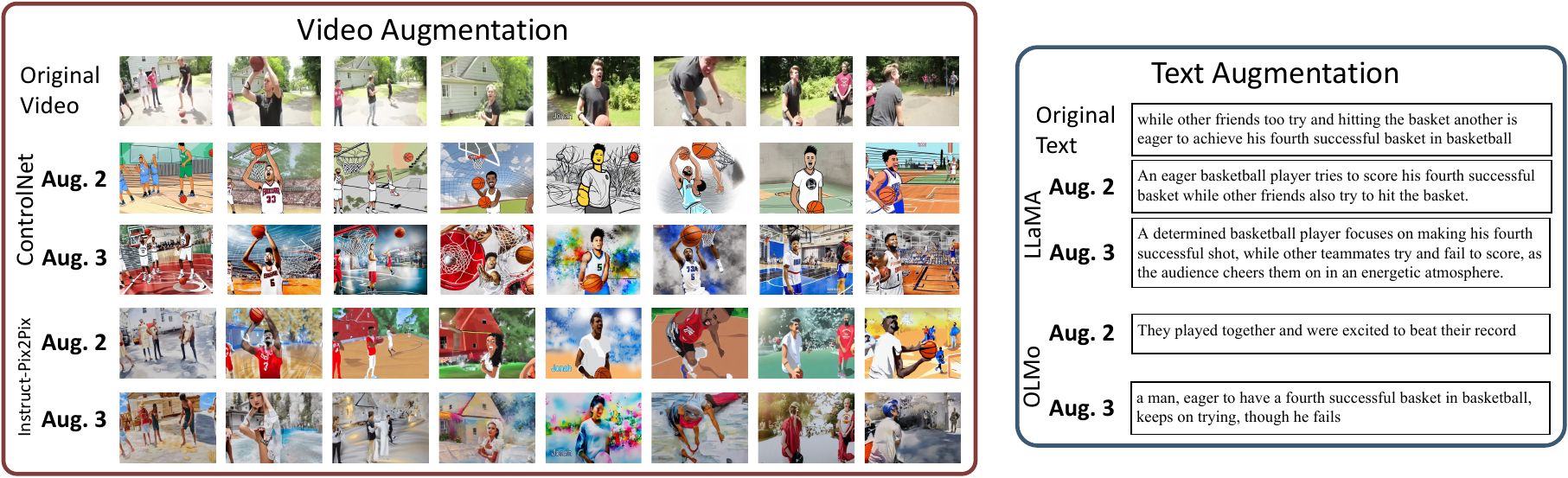}
\caption{
Qualitative examples of data generated by \ours. 
``Aug. 2'' and ``Aug. 3'' refer to augmentation by text paraphrasing and video stylization and augmentation by relevance enhancing. 
}\label{fig: qualitative full}
\end{figure*}

\textbf{MSVD}~\cite{chen-dolan-2011-collecting} contains 1,970 videos with a split of 1200, 100, and 670 as the train, validation, and test set, respectively.
The duration of videos varies from 1 to 62 seconds. 
Each video is paired with 40 English captions. 

\textbf{ActivityNet}~\cite{caba2015activitynet} is consisted of 20,000 Youtube videos with 100,000 densely annotated descriptions. 
For a fair comparison, following the previous setting~\cite{DBLP:journals/ijon/LuoJZCLDL22,DBLP:conf/eccv/Gabeur0AS20}, we concatenate all captions together as a paragraph to perform a video-paragraph retrieval task by concatenating all the descriptions of a video. 
Performances are reported on the ``val1'' split of the ActivityNet.

\subsection{Full Quantitative Results}

Due to the limitation of space, the results on video-to-text retrieval are presented in this technical appendix in \Cref{tab: activitynet full,tab: msvd full}. 
We still observe significant improvements broght by \ours\ compared with our baseline and previous SOTAs. 
Specifically, \ours\ improves the Recall@1 to 60.9 and 43.9 from 71.3 and 58.4 on MSVD and ActivityNet.

\subsection{Full Qualitative Results - Augmentation Examples}

To qualitatively validate the effectiveness of \ours, we present more augmentation examples on MSR-VTT in \Cref{fig: qualitative full}, respectively. 
It is notable that Instruct-Pix2Pix performs worse than ControlNet as Instruct-Pix2Pix puts a human head in the hands of a basketball player. 
Also, we observe that LLaMA consistently outperforms OLMo as it generates more details relevant to the original text. 
These qualitative results correspond to the quantitative results in \Cref{tab: appendix: different text and image gen}.

\begin{table}[t!]
\centering
\resizebox{\columnwidth}{!}{%
\begin{tabular}{ll|ccccc}
\toprule
                    \multirow{2}{*}{Methods}  & \multirow{2}{*}{Venue}&  \multicolumn{5}{c}{Video-to-Text Retrieval}           \\
                    &           & R@1$\uparrow$ & R@5$\uparrow$ & R@10$\uparrow$ & MdR$\downarrow$ & MnR$\downarrow$ \\
                               \midrule
CLIP4Clip & NeurComp'22 & 62.0 & 87.3 & 92.6 & 1.0 & 4.3    \\
CLIP2Video    &    Arxiv'21             & 58.7 & 85.6 & 91.6 & 1.0 & 4.3       \\
NCL    & EMNLP'22 &  69.6 & 89.9 & 95.4 & 1.0 & 3.3  \\
CenterCLIP  & SIGIR'22  & 57.9 & 83.6 & 90.5 & 1.0 & 5.2     \\
\midrule
X-CLIP    & ACM MM'22   & 60.9 & 87.8 & - &  - & 4.7         \\
\rowcolor{green!10}\ours & & \textbf{71.3} & \textbf{89.4} & \textbf{96.0} & \textbf{1.0} & \textbf{4.0}  \\
\bottomrule
\end{tabular}%
}
\caption{Video-to-Text retrieval results on MSVD. 
Best in \textbf{bold}.} 
\label{tab: msvd full}
\end{table}

\begin{table}[t!]
\centering
\resizebox{\columnwidth}{!}{%
\begin{tabular}{ll|ccccc}
\toprule
                    \multirow{2}{*}{Methods}  & \multirow{2}{*}{Venue}& \multicolumn{5}{c}{Video-to-Text Retrieval}           \\
                    &           & R@1$\uparrow$ & R@5$\uparrow$ & R@10$\uparrow$ & MdR$\downarrow$ & MnR$\downarrow$ \\
                               \midrule
CLIP4Clip & NeurComp'22 & 42.5 & 74.1 & 85.8 & 2.0 & 6.6 \\
VCM        & AAAI'22                     & 42.6 & 74.9 & - & 2.0 & 6.4         \\
NCL    & EMNLP'22 & 46.8 & 76.5 & 86.8 & 2.0 & 6.2      \\
Align\&Tell   & TMM     & 43.5 & 73.6 & - & 2.0 & -           \\
CenterCLIP  & SIGIR'22  & 44.5 & 75.3 & 86.0 & 2.0 & 6.7           \\
HBI & CVPR'23 &42.4 & 73.0 & 86.0 & 2.0 & 6.5 \\
UCoFiA & ICCV'23 &46.3 & 76.5 & - & - & - \\
\midrule
X-CLIP    & ACM MM'22          & 43.9 & 73.9 & - & - & 7.6         \\
\rowcolor{green!10}\ours   &  &   \textbf{58.4} & \textbf{85.2} & \textbf{88.7} & \textbf{1.0} & \textbf{5.7}   \\
\bottomrule
\end{tabular}%
}
\caption{Video-to-Text retrieval results on ActivityNet. Best in \textbf{bold}. 
}
\label{tab: activitynet full}
\end{table}

\begin{table*}[t!]
\centering
\resizebox{\textwidth}{!}{%
\begin{tabular}{c|ccccc|cccccc}
\toprule
\multirow{2}{*}{\# of Text}  & \multicolumn{5}{c|}{Text-to-Video Retrieval}                                          & \multicolumn{5}{c}{Video-to-Text Retrieval}                                                                                     \\
    & R@1$\uparrow$ & R@5$\uparrow$ & R@10$\uparrow$ & MdR$\downarrow$ & MnR$\downarrow$ & R@1$\uparrow$ & R@5$\uparrow$ & R@10$\uparrow$ & MdR$\downarrow$ & MnR $\downarrow$ \\\midrule
Baseline &    46.1          & {74.3}          & {83.1}           & 2.0             & {13.2}            & {46.8}          & 73.3          & {84.0}           & 2.0             & {9.1} \\
\midrule
\rowcolor{gray!10}\multicolumn{11}{c}{\textit{Augmentation 1: Simple Augmentation}}\\
1 & 50.8 & 76.7 & 84.6 & 1.0 & 11.1 & 53.6 & 76.6 & 85.2 & 1.0 & 8.7 \\
2 & 50.1 & 75.3 & 84.1 & 1.0 & 12.2 & 51.3 & 76.3 & 84.1 & 1.0 & 9.5 \\
3 & 51.3 & \textbf{77.0} & \textbf{85.5} & 1.0 & 11.3 & 52.9 & \textbf{78.1} & 85.0 & 1.0 & \textbf{8.3} \\
4 & 51.6 & 76.4 & 84.7 & 1.0 & 11.5 & 53.9 & 78.0 & 84.7 & 1.0 & \textbf{8.3} \\
5 & \textbf{52.0} & 76.5 & 84.7 & 1.0 & \textbf{11.0} & \textbf{51.2} & 76.5 & \textbf{85.3} & 1.0 & 10.0 \\
\midrule
\rowcolor{gray!10}\multicolumn{11}{c}{\textit{Augmentation 2: Augmentation by Text Paraphrasing and Video Stylization}}\\
1 & \textbf{56.1} & \textbf{81.3} & \textbf{89.4} & 1.0 & 9.7 & 52.4 & 77.3 & 84.9 & 1.0 & \textbf{8.0} \\
2 & 55.7 & 80.8 & 88.8 & 1.0 & \textbf{9.6} & 53.4 & 80.4 & 87.7 & 1.0 & 8.3 \\
3 & 56.0 & 80.5 & 88.7 & 1.0 & 10.0 & \textbf{55.2} & \textbf{81.4} & \textbf{89.3} & 1.0 & \textbf{8.0} \\
\midrule
\rowcolor{gray!10}\multicolumn{11}{c}{\textit{Augmentation 3: Augmentation by Relevance Enhancing}}\\
1 & 56.7 & 83.0 & 90.1 & 1.0 & 8.1 & 56.7 & 83.1 & 90.4 & 1.0 & 6.5 \\
2 & 60.1 & 83.1 & 90.1 & 1.0 & {5.9} & 57.7 & {84.2} & {91.6} & 1.0 & {6.2} \\
3 & {60.4} & 85.3 & 91.0 & 1.0 & \textbf{5.8} & {60.3} & 84.1 & 91.3 & 1.0 & {6.6}\\
4 & 59.2 & 84.0 & {91.1} & 1.0 & 7.1 & 57.3 & 83.8 & 90.3 & 1.0 & 7.4\\
\rowcolor{green!10}5 & \textbf{60.8}   & \textbf{84.5}   & \textbf{91.4}   & 1.0    & \textbf{5.8}    & \textbf{60.6}   & \textbf{85.2}   & \textbf{92.5}   & 1.0   & \textbf{5.9} \\
                          \bottomrule
\end{tabular}%
    }
\caption{
Full retrieval performance with different numbers of augmented captions using three augmentation methods on MSR-VTT. 
Best in \textbf{bold} and the results on video-to-text retrieval are deferred to Appendix due to the space limitation. 
}
\label{tab: ablation: number of text full}
\end{table*}

\begin{table*}[t!]
\centering
\resizebox{\textwidth}{!}{%
\begin{tabular}{c|ccccc|cccccc}
\toprule
\multirow{2}{*}{\# of Video}  & \multicolumn{5}{c|}{Text-to-Video Retrieval}                                          & \multicolumn{5}{c}{Video-to-Text Retrieval}                                                                                     \\
    & R@1$\uparrow$ & R@5$\uparrow$ & R@10$\uparrow$ & MdR$\downarrow$ & MnR$\downarrow$ & R@1$\uparrow$ & R@5$\uparrow$ & R@10$\uparrow$ & MdR$\downarrow$ & MnR $\downarrow$ \\\midrule
Baseline &    46.1          & {74.3}          & {83.1}           & 2.0             & {13.2}            & {46.8}          & 73.3          & {84.0}           & 2.0             & {9.1} \\
\midrule
\rowcolor{gray!10}\multicolumn{11}{c}{\textit{Augmentation 1: Simple Augmentation}}\\
1 & \textbf{50.4} & 75.8 & 84.9 & 1.0 & 11.2 & \textbf{52.1} & 76.9 & 85.0 & 1.0 & 9.1 \\
2 & 48.7 & 75.6 & 84.3 & 2.0 & 11.4 & 50.7 & 76.6 & 84.2 & 1.0 & 8.9 \\
3 & 50.0 & 75.0 & 84.2 & 1.5 & 12.4 & 48.9 & 76.5 & 84.4 & 2.0 & 9.6 \\
4 & 49.8 & 74.4 & 82.8 & 2.0 & 11.6 & 49.9 & 74.1 & 84.3 & 2.0 & 10.3 \\
5 & \textbf{50.4} & \textbf{77.2} & \textbf{86.0} & 1.0 & \textbf{10.7} & 51.6 & \textbf{77.6} & \textbf{85.4 }& 1.0 & \textbf{8.5} \\
\midrule
\rowcolor{gray!10}\multicolumn{11}{c}{\textit{Augmentation 2: Augmentation by Text Paraphrasing and Video Stylization}}\\
1 & 53.8 & \textbf{80.7} & 88.6 & 1.0 & 9.9 & 55.6 & 79.6 & 88.2 & 1.0 & 7.7 \\
2 & 54.4 & 79.3 & 87.4 & 1.0 & 10.1 & \textbf{56.4} & 80.3 & 87.9 & 1.0 & 7.2\\
3 & \textbf{54.7} & 80.3 & \textbf{88.9} & 1.0 & \textbf{9.3} & 55.3 & \textbf{80.8} & 87.9 & 1.0 & \textbf{6.8} \\
\midrule
\rowcolor{gray!10}\multicolumn{11}{c}{\textit{Augmentation 3: Augmentation by Relevance Enhancing}}\\
1 & 60.4 & 84.4 & \textbf{91.4} & 1.0 & 6.2 & 59.3 & 84.8 & 91.5 & 1.0 & 6.5\\
2 & 60.0 & 84.0 & 91.2 & 1.0 & 6.4 & 58.8 & \textbf{85.9} & 91.4 & 1.0 & 7.0 \\
\rowcolor{green!10}3 & \textbf{60.8}   & \textbf{84.5}   & \textbf{91.4}   & 1.0    & \textbf{5.8}    & \textbf{60.6}   & {85.2}   & \textbf{92.5}   & 1.0   & \textbf{5.9} \\
\bottomrule
\end{tabular}%
    }
\caption{
Full retrieval performance with different numbers of generated videos using three augmentation methods on MSR-VTT. 
Best in \textbf{bold}. 
}
\label{tab: ablation: number of video full}
\end{table*}

\begin{table*}[t!]
\centering
\resizebox{\textwidth}{!}{%
\begin{tabular}{lc|ccccc|ccccc}
\toprule
 \multirow{2}{*}{Methods}  & \multirow{2}{*}{LLMs}          & \multicolumn{5}{c|}{Text-to-Video Retrieval}                                         & \multicolumn{5}{c}{Video-to-Text Retrieval}                                          \\
        &     & R@1$\uparrow$ & R@5$\uparrow$ & R@10$\uparrow$ & MdR$\downarrow$ & MnR$\downarrow$ & R@1$\uparrow$ & R@5$\uparrow$ & R@10$\uparrow$ & MdR$\downarrow$ & MnR$\downarrow$ \\
\midrule
Baseline   & -  & 46.1          & {74.3}          & {83.1}           & 2.0             & {13.2}            & {46.8}          & 73.3          & {84.0}           & 2.0             & {9.1}             \\
\midrule
\rowcolor{gray!10}\multicolumn{12}{c}{\textit{Augmentation 2: Augmentation by Text Paraphrasing and Video Stylization}}\\
 \multirow{2}{*}{\ours}  & LLaMA-2-7b-chat-hf & 54.7 & 80.3 & 88.9 & 1.0 & 9.3 & 55.3 & 80.8 & 87.9 & 1.0 & 6.8 \\
   &  OLMo-7b &     53.2 &  77.9 & 86.3 & 1.0 & 9.8 & 53.7 & 77.4 & 87.2 & 1.0 & 7.5        \\
\midrule
\rowcolor{gray!10}\multicolumn{12}{c}{\textit{Augmentation 3: Augmentation by Relevance Enhaching}}\\
\rowcolor{green!10} \cellcolor{white!10}{\multirow{2}{*}{\ours}}  &   LLaMA-2-7b-chat-hf & \textbf{60.8}   & \textbf{84.5}   & \textbf{91.4}   & \textbf{1.0}    & \textbf{5.8}    & \textbf{60.6}   & \textbf{85.2}   & \textbf{92.5}   & \textbf{1.0}    & \textbf{5.9}             \\
   &  OLMo-7b &       58.7 & 81.2 & 89.5 & 1.0 & 7.5 & 59.2 & 83.7 & 90.3 & 1.0 & 7.1       \\
\bottomrule
\end{tabular}%
}
\caption{
Full retrieval results on MSR-VTT using different LLMs for generating paraphrased captions. 
Best in \textbf{bold}. 
}
\label{tab: appendix: different llms full}
\end{table*}

\begin{table*}[t!]
\centering
\resizebox{\textwidth}{!}{%
\begin{tabular}{lc|ccccc|ccccc}
\toprule
 \multirow{2}{*}{Methods}  & \multirow{2}{*}{VGMs}          & \multicolumn{5}{c|}{Text-to-Video Retrieval}                                        & \multicolumn{5}{c}{Video-to-Text Retrieval}                                        \\
        &     & R@1$\uparrow$ & R@5$\uparrow$ & R@10$\uparrow$ & MdR$\downarrow$ & MnR$\downarrow$ & R@1$\uparrow$ & R@5$\uparrow$ & R@10$\uparrow$ & MdR$\downarrow$ & MnR$\downarrow$ \\
\midrule
Baseline   & -  & 46.1          & {74.3}          & {83.1}           & 2.0             & {13.2}            & {46.8}          & 73.3          & {84.0}           & 2.0             & {9.1}             \\
\midrule
\rowcolor{gray!10}\multicolumn{12}{c}{\textit{Augmentation 2: Augmentation by Text Paraphrasing and Video Stylization}}\\
 \multirow{2}{*}{\ours} & ControlNet  &  54.7 & 80.3 & 88.9 & 1.0 & 9.3 & 55.3 & 80.8 & 87.9 & 1.0 & 6.8 \\
   &  instruct-pix2pix &     56.7 & 83.0 & 90.1 & 1.0 & 8.1 & 56.7 & 83.1 & 90.4 & 1.0 & 6.5         \\
\midrule
\rowcolor{gray!10}\multicolumn{12}{c}{\textit{Augmentation 3: Augmentation by Relevance Enhaching}}\\
\rowcolor{green!10} \cellcolor{white!10}{\multirow{2}{*}{\ours}}  &   ControlNet & \textbf{60.8}   & \textbf{84.5}   & \textbf{91.4}   & \textbf{1.0}    & \textbf{5.8}    & \textbf{60.6}   & \textbf{85.2}   & \textbf{92.5}   & \textbf{1.0}    & \textbf{5.9}             \\
   &  instruct-pix2pix &      57.8 & 82.9 & 90.2 & 1.0 & 7.9 & 58.2 & 81.4 & 87.8 & 1.0 & 8.0        \\
\bottomrule
\end{tabular}%
}
\caption{
Full retrieval results on MSR-VTT using different image generation methods for generating stylized video frames. 
Best in \textbf{bold}. 
}
\label{tab: appendix: different image gen}
\end{table*}

\subsection{Full Ablation Studies}

\noindent
\textbf{Number of paraphrased texts.} 
\Cref{tab: ablation: number of text,tab: ablation: number of text full} offers an insightful look into the effects of varying the number of augmented text on the retrieval performance in the MSR-VTT dataset, utilizing \ours\ with X-CLIP (ViT-B/32) as the base model. 
With simple augmentation, a gradual increase in performance is seen with the number of texts, peaking at 4 texts for text-to-video retrieval with a Recall@1 of 51.6, and similarly for video-to-text retrieval at 53.9. 
This suggests that self-generated augmentations contribute positively to the model's understanding and retrieval capabilities.
Moving to augmentation by text paraphrasing and video stylization, the performance leaps further, highlighting the value of leveraging external, sophisticated models to enrich the dataset. 
The best results are achieved with 3 texts, indicating an optimal balance between augmentation quantity and retrieval efficacy.
The last augmentation method, augmentation by relevance enhancing, showcases the most significant performance boosts, especially with 5 texts, achieving the highest Recall@1 of 60.8 for text-to-video and 60.6 for video-to-text retrieval, along with the best Recall@5, Recall@10, and the lowest MdR and MnR scores. 
This illustrates the power of relevance-based augmentation in dramatically enhancing retrieval accuracy by introducing more diverse and complex semantic representations into the training process.

\noindent
\textbf{Number of generated videos.} 
\Cref{tab: ablation: number of video,tab: ablation: number of video full} presents an ablation study on the impact of different numbers of generated videos. 
For simple augmentation, we notice considerable improvements, particularly with 5 videos, achieving a Text-to-Video Retrieval (T2V) R@1 of 50.4, and a Video-to-Text Retrieval (V2T) R@1 of 51.6. 
When moving to augmentation by text paraphrasing and video stylization, it further elevates the performance, with the best results observed when using 3 videos, where T2V R@1 reaches 54.7 and V2T R@1 peaks at 56.4. 
This suggests that leveraging foundation models for augmentation can significantly impact retrieval effectiveness, likely due to the richer, more diverse semantic representations introduced.
The last strategy, augmentation by relevance enhancing, achieves the highest T2V R@1 of 60.8 and V2T R@1 of 60.6, alongside the best R@5 and R@10 scores, underscoring the efficacy of relevance-based augmentation in capturing nuanced, diverse semantic content, thus markedly improving retrieval precision. 
The progressive increase in retrieval performance across the augmentation strategies, especially with the augmentation by relevance enhancing, highlights the potential of sophisticated, creative augmentation techniques in enhancing the capabilities of text-video retrieval.

\noindent\textbf{Choice of LLM for paraphrasing.} 
To understand how LLMs impact retrieval performance, we also use OLMo~\cite{groeneveld2024olmo} for generating paraphrased captions. 
The results are shown in \Cref{tab: appendix: different text and image gen,tab: appendix: different llms full}. 
The data showcases a notable improvement when using LLMs over the baseline, with LLaMA achieving the highest performance across all metrics, underscoring its superiority in understanding and generating nuanced paraphrased captions that significantly benefit retrieval accuracy. 
Specifically, LLaMA demonstrates exceptional capability in both augmentation strategies but particularly excels in Augmentation by relevance enhancing, where it achieves the best recall rates (60.8 in Text-to-Video and 60.6 in Video-to-Text) and the lowest rank metrics (MdR and MnR at 1.0 and 5.8/5.9, respectively). 
These findings underscore the importance of selecting appropriate LLMs for paraphrase generation in multimedia retrieval tasks, highlighting how advanced models like LLaMA can effectively bridge semantic gaps between text and video content to improve retrieval outcomes.

\noindent \textbf{Choice of image generation methods.} 
To understand how image generation methods impact retrieval performance, we also use Instruct-Pix2Pix~\cite{brooks2022instructpix2pix} for generating video frames. 
The results are shown in \Cref{tab: appendix: different image gen}. 
It underscores the superiority of the ControlNet method in the augmentation by relevance enhancing category, marking it with the highest recall scores (60.8 for Text-to-Video and 60.6 for Video-to-Text) and the lowest ranks (MdR and MnR of 1.0 and 5.8/5.9, respectively), significantly outperforming the baseline and showcasing the potential of leveraging relevance-bsaed techniques in enhancing video-text retrieval tasks. 
It also highlights the pivotal role of advanced image generation methods in improving the semantic alignment between video and text, offering promising directions for future research in multimedia retrieval.

\end{document}